\documentclass{article}
\usepackage[preprint]{spconf}
\usepackage{amsmath,amssymb,graphicx,xfrac,multirow,url}
\usepackage[ruled]{algorithm2e}
\usepackage{algcompatible,lipsum,url,tikz}
\newsavebox{\algleft}
\newsavebox{\algcenter}
\newsavebox{\algright}

\title{Tensor Reordering for CNN Compression}

\name{Matej Ulicny$^{\star}$, Vladimir A. Krylov$^{\diamond}$ \& Rozenn Dahyot$^{\star}$\thanks{This research was supported by the ADAPT Centre for Digital Content Technology funded under the SFI Research Centres Programme (Grant 13/RC/2106) and co-funded under the European Regional Development Fund.}}
\address{
$^{\star}$School of Computer Science \& Statistics, ADAPT Research Centre, Trinity College Dublin, Ireland\\
$^{\diamond}$School of Mathematical Sciences, ADAPT Research Centre, Dublin City University,  Ireland}

\begin{document}
\ninept

\maketitle

\begin{abstract}
We show how parameter redundancy in Convolutional Neural Network (CNN) filters can be effectively reduced by pruning in spectral domain. Specifically, the representation extracted via Discrete Cosine Transform (DCT) is more conducive for pruning than the original space. By relying on a combination of weight tensor reshaping and reordering we  achieve high levels of layer compression with just minor accuracy loss. 
Our approach is applied to compress pretrained CNNs and we show that minor additional fine-tuning allows our method to recover the original model performance after a significant parameter reduction. We validate our approach on ResNet-50 and MobileNet-V2 architectures for ImageNet classification task.
\end{abstract}
\begin{keywords}
Compression, Pruning, Reordering, DCT, CNN 
\end{keywords}

\section{Introduction}
\label{sec:intro}

A wide variety of methods have been proposed in the recent years to perform structured (i.e., removing entire filters of blocks) and unstructured (i.e., suppressing individual weights) pruning of CNNs. Such studies are central to addressing the design of memory- and computationally-efficient implementations of the state-of-the-art CNNs. It has been pointed out in multiple works~\cite{he2018soft,gale2019state,mlsys2020_73} that standard CNN architectures are typically highly over-parameterized to the point that pruning of some 5-20\% of parameters may often result in networks performing better than the original architectures. This impact can be associated with better optimization and regularization. Iterative pruning~\cite{Liebenwein2020Provable,frankle2019linear,he2018amc} and access to substantial fine-tuning~\cite{luo2017thinet,lin2020channel,you2019gate} or even re-training from scratch~\cite{frankle2019linear,renda2019comparing} allows many techniques to arrive at the (nearly) original levels of performance 
at the cost of additional 30\% to 300\% of the original amount of training.
This is achieved after dropping between 30 and 70\% of the weight parameters in general-purpose architectures like VGG~\cite{simonyan2014very} and ResNets~\cite{he2016deep}, and 5-20\% on architectures designed to run on devices with limited resources, like MobileNets~\cite{sandler2018mobilenetv2}. 

Most existing pruning methods aim at reducing the computational complexity while training or deploying  CNNs, 
 and often consider the decrease of parameter count (and hence memory footprint) of the pruned architectures as secondary~\cite{he2018soft,mlsys2020_73}. 
 Contrary to these approaches we focus here specifically on the parameter footprint aspect and consider scenarios when very limited or no retraining budget may be available to fine-tune the compressed models. The motivation for such goal is two-fold. Firstly, we consider the storage implications (and overheads related to compressing the original models) for models both in trained architecture repositories, i.e. model zoos and mobile devices. Secondly, by reducing the trainable parameter count in the models we expect to retain the general modeling capacity and increase the degree of regularization for the compressed architecture optimization thus reducing the impact of over-fitting. 
Summarily, we propose a non-iterative technique and resort to a strictly limited amount of fine-tuning to compensate for the impact of pruning: we let all our compressed networks train for a single epoch after compression.

For the general state-of-the-art review on CNN pruning we refer the readers to more comprehensive recent overviews conducted in~\cite{mlsys2020_73,wang2018packing}.
Several recent studies~\cite{wang2018packing,dubey2018coreset} have also investigated full-cycle compression including on top of CNN-specific elements the general instruments like quantization and Huffman coding thus achieving the reduction of memory footprint of 10-50 times for AlexNet, VGG16, ResNet-50. In our work we do not investigate specifically the memory footprint but rather the parameter count, assuming that the stronger parameter reduction generally leads to better compressibility.

\begin{figure}[!h]
\centering
\includegraphics[width=.91\linewidth]{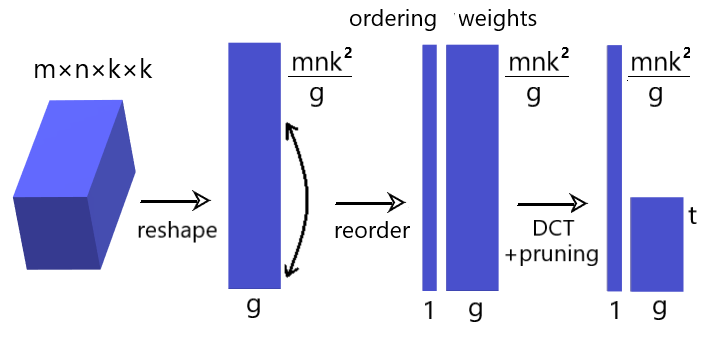}
\vspace{-1.5\baselineskip}
\caption{Tensor reordering and DCT compression.}
\label{fig:diagram}
\end{figure} 

A number of works have explored links between CNNs with wavelet decomposition~\cite{Bruna13,Oyallon18}, and demonstrated how efficient compression can be achieved by casting spatial CNN filters into wavelet bases such as DCT~\cite{Ulicny19b} or Gaussian derivatives~\cite{Kobayashi18}. In our work we aim to reduce  redundancies due to correlation between the filters in the same network layer, and to this end we use spectral representation where we perform the weights compression.
We demonstrate that DCT can be used to compress more efficiently if deployed along the channel dimensions (Sec.~\ref{sec:DCT}) compared to the more common spatial filter compression.  Our approach relies on tensor reordering to maximise the representation capacity of DCT encoding such that efficient compression can be achieved by suppressing high frequencies, see Fig.~\ref{fig:diagram}.
Frequency based pruning~\cite{Liu18frequency} has been previously used to compress convolutional filters based on their spatial correlation, however without attempting to compress layers with $1\times1$ kernels or fully connected layers; Liu et al approach is dynamic and prunes DCT coefficients based on their magnitude. In contrast our static approach is truncating coefficients only by their frequency and does not require extensive fine-tuning.

The implementation of our proposed compression technique\footnote{PyTorch implementation is available at \url{https://github.com/matej-ulicny/reorder-cnn-compression}.} is assessed in terms of accuracy vs. parameter count against state-of-the-art pruning approaches including~\cite{he2018soft,gale2019state,Liebenwein2020Provable,luo2017thinet,lin2020channel,renda2019comparing,dubey2018coreset,lin2019towards,zhuang2018discrimination,wang2020}.
We present strong results with our novel CNN compression applied to ResNet-50~\cite{he2016deep} and MobileNet-V2~\cite{sandler2018mobilenetv2} architectures for image classification task on ImageNet ILSVRC dataset~\cite{russakovsky2015imagenet} (Sec.~\ref{secexp}). 
The selection of these two state-of-the-art architectures on a large challenging dataset provides a suitable validation for the proposed compression technique.
We achieve the original ResNet-50 performance with 32\% of the original parameter footprint (with 6\% of trainable parameters) with just one epoch of post-compression fine-tuning. Furthermore, our compressed models without fine-tuning achieve competitive results on  par with recent compression methods involving ample fine-tuning. On MobileNet-V2 we achieve almost 50\% parameter reduction with just 0.3\% drop in top-1 accuracy.

\section{Tensor compression via DCT}
\label{sec:DCT}

The vast majority of parameters representing the convolutional neural network are concentrated in convolutional filters or weight matrices of the fully connected layers.
A set of filters representing a convolutional layer is typically expressed as a 4-dimensional tensor $\mathbf{w}\in\mathbb{R}^{m\times n\times k\times k}$ and is used to convolve each of the $n$ input feature maps with its own $k\times k$ filter, spanning $m$ output feature maps:
\begin{equation} \label{eq:convolution}
    \mathbf{y}_j = \sum_{i=0}^{n-1} \mathbf{x}_i * \mathbf{w}_{j,i}.
\end{equation}
Many of the layers in recent architectures are so called resampling layers, whose filters have $1\times 1$ spatial extent ($k=1$), i.e., they merely reweight existing features. A fully-connected layer can be treated as a special case of convolutional layer with $k=1$ that resamples features consisting of a single scalar.
Our aim is to compress the weights of a layer by deploying one-dimensional DCT transform. The columns of the transform matrix $C\in\mathbb{R}^{n\times n}$ represent the DCT basis functions, with elements $C\left(a,u\right)$:
\begin{equation} \label{eq:dct_matrix_cell}
\begin{aligned}
C&\left(a,u\right) = \sqrt{\frac{\alpha\left(u\right)}{n}} \text{cos}\left[ \frac{\pi}{n} \left( a + \frac{1}{2} \right) u \right]\\
\end{aligned}
\end{equation}
where $\alpha\left(u=0\right) = 1, \alpha\left(u>0\right) = 2$.
Given a weight tensor $\mathbf{w}\in\mathbb{R}^{m\times n \times 1 \times 1}$, the output channel $j$ ($j=0,...,m-1$) is calculated from the input using a weight vector $\mathbf{w}_j\in\mathbb{R}^n$. In $\mathbf{w}_j$ each element represents the weight of an input channel. The DCT matrix $\mathrm{C}$ is orthogonal, and therefore $\mathbf{w}_j = \mathrm{C}\mathrm{C}^T\mathbf{w}_j$.
The weight vector $\mathbf{z}_j$ in the DCT domain $\mathbf{z}_j=\mathrm{C}^T\mathbf{w}_j$ also has $n$ coefficients. We apply compression to each vector $\mathbf{w}_j$ by retaining its $t < n$ lowest frequencies.
The number of DCT basis functions $t = \left\lfloor \frac{n}{r} \right\rfloor$ used to transform the weights is controlled by a hyperparameter $r$ that represents the fraction of the complete basis set used after compression and is referred to as compression rate in the following.
The reduced transformation matrix $\mathrm{C}_t\in\mathbb{R}^{n\times t}$ will produce reduced coefficient representation $\mathbf{z}^t_j=\mathrm{C}_t^T\mathbf{w}_j$. The tensor $\mathbf{z}^t$ is representing our compressed layer in DCT domain, which is then used at runtime to reconstruct an approximation $\mathbf{\tilde{w}}$ of the original weight tensor $\mathbf{w}$ via $\mathbf{\tilde{w}}_j = \mathrm{C}_t\mathbf{z}_j^t = \mathrm{C}_t\mathrm{C}_t^T\mathbf{w}_j.$

It is well-known that concise Fourier and DCT representations are achieved on smooth signals, whereby  discontinuities require a substantial number of additional coefficients in the representation. Similarly,
compression based on frequency truncation assumes smooth transitions between neighboring elements. Unlike parameters at neighboring spatial locations~\cite{Jacobsen16}, the weights corresponding to neighboring channels of a CNN are not expected to be correlated. From  equation~(\ref{eq:convolution}) it is clear however that the exact channel position does not matter as long as the filter responses are summed. For this reason, before applying DCT to such channels, we boost coherence between the input channels by reordering them. Our expectation is that the more similar the neighboring samples are, the fewer basis functions are required to encode the signal accurately.

Considering a 2D weight matrix $\mathbf{w}\in\mathbb{R}^{m\times n}$, the 1D DCT transform is to be applied $m$ times for every row-vector $\mathbf{w}_{j,\cdot}$, $j\in\left\{0..m-1\right\}$ of size $n$. To boost the smoothness of $\mathbf{w}$, we deploy the ordering procedure outlined in Alg.~\ref{alg:reorder}. Specifically, we reorder the weight matrix $\mathbf{w}$ along the second dimension of size $n$. The new sequence starts with the column-vector $\mathbf{w}_{\cdot,i}$, $i\in\left\{0..n-1\right\}$ with the largest magnitude. Each following vector is chosen to have the smallest distance $\mathcal{D}$ to its predecessor.

\begin{algorithm}[!h]
 \KwIn{$\mathbf{w}\in\mathbb{R}^{m\times n}$ (weight matrix)}
 $\mathbf{s} \in \mathbb{N}^{n}$;\\
 $\mathbf{s}_0 \leftarrow \arg\max_{i\in \left\{0..n-1\right\}}\left\Vert \mathbf{w}_{\cdot,i}\right\Vert_2$;\\
 $\mathbf{p} \leftarrow \left\{0..n-1\right\} \backslash \left\{\mathbf{s}_0\right\}$; \\
 \For{$j \in \left\{ 1..n-1 \right\}$}{
   $\mathbf{s}_j \leftarrow \arg\min_{i\in p} \mathcal{D}\left( \mathbf{w}_{\cdot,\mathbf{s}_{j-1}},\mathbf{w}_{\cdot,i}\right)$;\\
   $\mathbf{p} \leftarrow \mathbf{p} \backslash \left\{\mathbf{s}_j\right\}$; \\
 }
 \KwOut{$\mathbf{s}$ (ordering)}
 \caption{Weight matrix reordering} 
 \label{alg:reorder}
\end{algorithm}

In case of output dimension $m=1$ this procedure establishes the standard ordering from the largest to the smallest.
The purpose of the proposed ordering is to boost smoothness between columns of the reordered tensor. This however is low for large values of the first dimension $m$, corresponding to the size of vectors being sorted. The natural approach to decrease $m$ is to reshape $\mathbf{w}$ to $\mathbf{w}'\in\mathbb{R}^{g\times mn/g}$ such that the first dimension is $g < m$ - the number of groups we reshape the layer into. The reshaping operation is defined as $\mathcal{R}_{{m\times n} \rightarrow {g\times mn/g}}\mathbf{w}$ and can be formally described as inverse vectorization vec$_{g,mn/g}^{-1}\circ$vec $\mathbf{w}$. This transformation allows higher smoothness of transitions inside the reordered $\mathbf{w}$ for lower values of $g$. Alg.~\ref{alg:compression} describes the compression procedure with reshaping, applied to arbitrary square kernel of size $k$. Fig.~\ref{fig:diagram} shows that reordering the weight tensor requires storing the ordering (index vector) to reverse this manipulation. The size of this vector is $mnk^2/g$, thus small values of $g$ span a long index vector $\textbf{s}$. This means that attaining higher levels of smoothness requires storing more extra information.

\begin{algorithm}[!h]
 \KwIn{$\mathbf{w}\in\mathbb{R}^{m\times n\times k \times k}$ (weight matrix)}
 $\mathbf{w}' \leftarrow \mathcal{R}_{m\times n\times k\times k\rightarrow g\times mnk^2/g}\mathbf{w}$; (reshape)\\
 $\mathbf{s} \leftarrow \text{Alg.~\ref{alg:reorder}}\left( \mathbf{w}' \right)$;\\
 $\mathbf{w}' \leftarrow \mathbf{w'}_{\cdot,\mathbf{s}}$; (ordering)\\
 $\mathbf{z}^t\in\mathbb{R}^{g\times t}$;\\
 \For{$j \in \left\{ 0..g-1 \right\}$}{
   $\mathbf{z}_j^t \leftarrow C_t^T\mathbf{w}'_j$;\\
 }
 \KwOut{$\mathbf{z}^t, \mathbf{s}$ (DCT weights, ordering)}
 \caption{Weight matrix compression} 
 \label{alg:compression}
\end{algorithm}

The decompression of the original weight tensor into $\mathbf{\tilde{w}}$ for inference and training consists of applying the inverse DCT (iDCT) to the stored coefficients, reordering the vectors to their original position and reshaping the weight tensor to its original shape, see Alg.~\ref{alg:decompression}.
The procedure consists of differentiable operations and allows gradient backpropagation. Our reordering, permutation and the inverse DCT transformation generate some additional computations on top of convolutions in comparison to standard structured pruning approaches.
While small values of $g$ increase complexity of the iDCT, we employ FFT algorithm having complexity $\mathcal{O}\left(mnk^2 \log{\frac{mnk^2}{g}}\right)$ that has only a limited impact on the total number of operations.
It is also important to note that the post-compression parameter footprint created by our approach consists of trainable parameters and order parameters $\mathbf{s}$. The latter are not changing throughout the fine-tuning phase. The fraction of trainable parameters can be between 15\% (low $g$) to 90\% (high $g$) of the total.

\begin{algorithm}[!h]
   \KwIn{$\mathbf{z}^t\in\mathbb{R}^{g\times t}$ (DCT weights)}
   {\phantom{+++++.}$\mathbf{s} \in \mathbb{N}^{mnk^2/g}$ (ordering)}\\
 $\mathbf{w}'\in\mathbb{R}^{g\times mnk^2/g}$;\\
 \For{$j \in \left\{ 0..g-1 \right\}$}{
   $\mathbf{w}'_j \leftarrow \mathrm{C}_t\ \mathbf{z}^t_j$; (inverse DCT)\\
 }
 $\mathbf{w}' \leftarrow \mathbf{w'}_{\cdot,\mathbf{s}^{-1}}$; (inverse ordering)\\
 $\mathbf{\tilde{w}} \leftarrow \mathcal{R}_{g\times mnk^2/g \rightarrow m\times n\times k\times k}\mathbf{w}'$; (reshape)\\
 \KwOut{$\mathbf{\tilde{w}}$ (decompressed weights)}
 \caption{Weight matrix decompression} 
 \label{alg:decompression}
\end{algorithm}

\section{Progressive compression}  
\label{sec:progr}

Compressing shallower layers increases classification error more than compressing deeper layers by the same ratio. These `hard-to-compress' layers have also substantially fewer parameters and compressing them with high ratios brings almost no change in model size but leads to a substantial drop in network performance. It is expected that after sorting, larger weight matrices produce less sparse sequence with regular structure that can be approximated with proportionally fewer basis functions. We tackle the performance/size trade-off by deriving compression parameters from the size of the layer: strategy \textit{progressive-r} denotes adjusting compression ratio $r$ with fixed $g$ throughout the network, and strategy \textit{progressive-g}  optimizes $g$  with fixed $r$. For strategy \textit{progressive-r}, a user provided hyperparameter $r'$ is used as a ratio increase $r_{l_1}=1+r'$ for a reference layer $l_1$ that we chose to be the smallest compressed layer. Let $p_{l_i}=m_{l_i}n_{l_i}k_{l_i}^2$ be the number of parameters of the weight tensor of layer $l_i$. The compression ratio $r_{l_i}$ for any layer $l_i$ is then 
\begin{equation}
  r_{l_i}=1+r'\frac{\sqrt{p_{l_i}}}{\sqrt{p_{l_1}}}.
\end{equation}
The proportion of square roots of parameter counts scales linearly with the number of channels. The strategy \textit{progressive-g} determines $g_{l_i}$ as the closest exponent of 2 that is lower or equal to the weight size proportion:
\begin{equation}
  g_{l_{i}}=\max{\left(2, 2^{\left \lfloor \log_2{\frac{\sqrt{p_{l_i}}}{\sqrt{p_{l_1}}}} \right \rfloor}\right)}.
\end{equation}
Finally  the  compression strategy \textit{uniform} propose to use the same compression rate $r$ and number of groups $g$  for all layers.


\section{Experiments}
\label{secexp}

To evaluate the proposed compression scheme outlined in Alg.~\ref{alg:compression} we perform experiments with recent ResNet-50 and MobileNet-V2 models trained on ImageNet-1k image classification dataset. We investigate the proposed compression in term of reconstruction error (Sec.~\ref{secexp:resample}) and overall accuracy on validation set (Sec.~\ref{secexp:combine}). 

\begin{figure*}[t]
\begin{center}
\begin{minipage}[b]{.495\linewidth}
  \centering
  \centerline{\includegraphics[width=\linewidth]{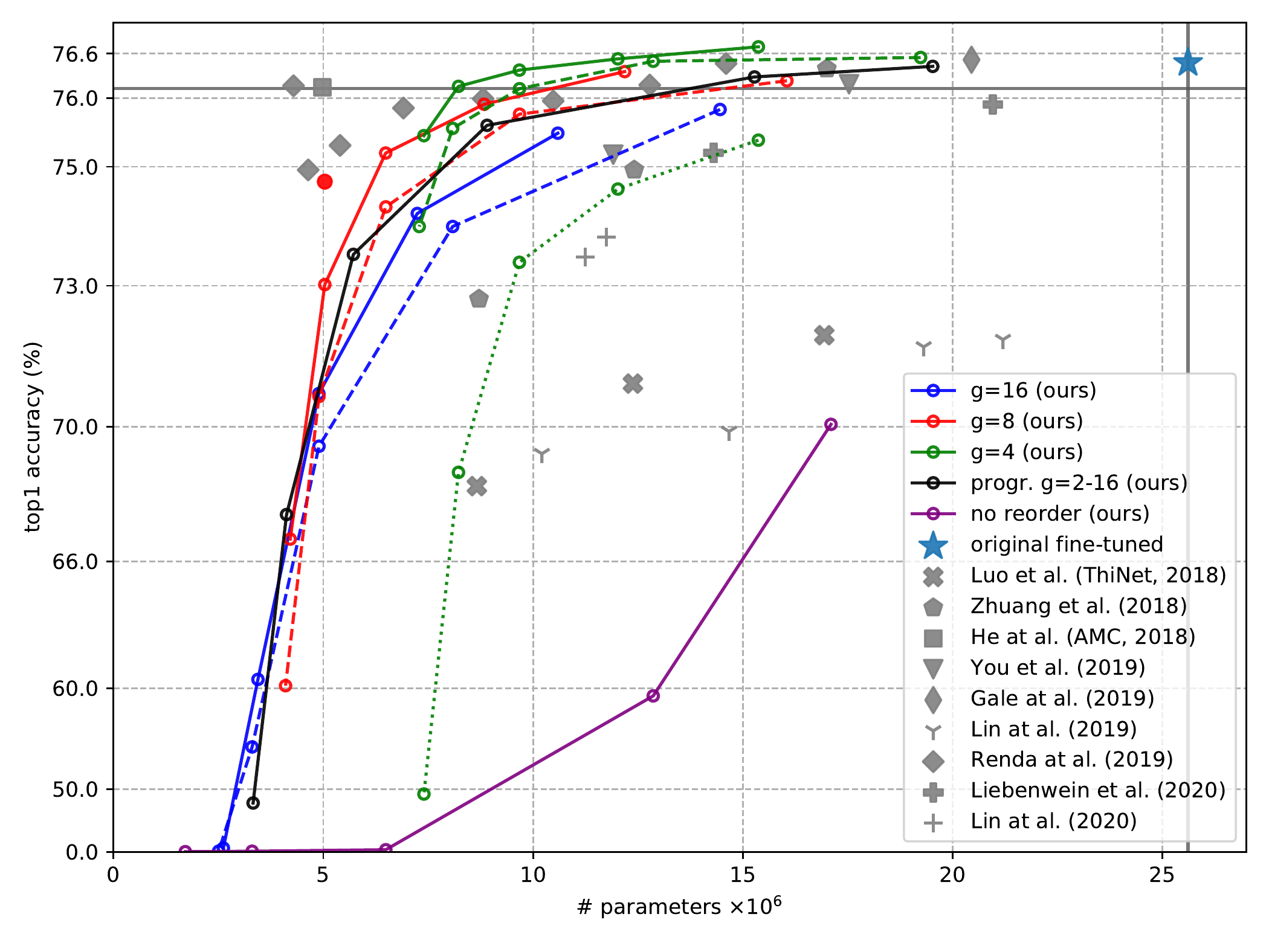}}
  \vspace{-.2cm}
  \centerline{ResNet-50 compression on Imagenet}
\end{minipage}
\hfill
\begin{minipage}[b]{0.495\linewidth}
  \centering
  \centerline{\includegraphics[width=\linewidth]{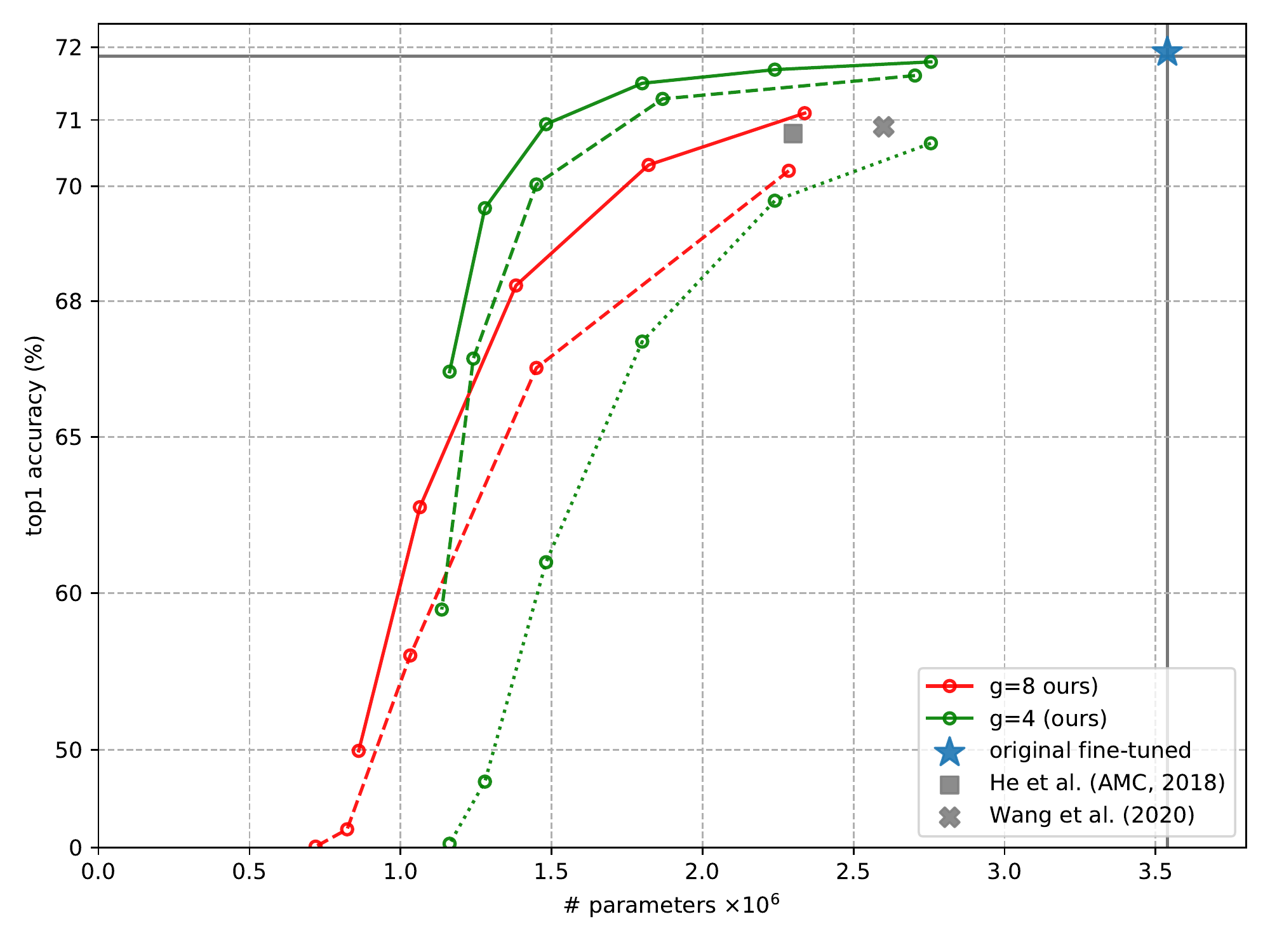}}
  \vspace{-.2cm}
  \centerline{MobileNet-V2 compression on Imagenet}
\end{minipage}
\vspace{-1.25\baselineskip}
\end{center}
   \caption{Top-1 accuracy vs. parameter count on ImageNet validation set after one epoch of fine-tuning for (a) ResNet-50 and (b) MobileNet-V2 architectures: progressive-$r$ compression strategy (solid) clearly outperforms uniform (dashed). The model without fine-tuning (dotted) outperforms some of the related works~\cite{he2018soft,gale2019state,Liebenwein2020Provable,luo2017thinet,lin2020channel,renda2019comparing,dubey2018coreset,lin2019towards,zhuang2018discrimination,wang2020}. The red dot in (a) shows performance after 20 epoch fine-tuning. Grey vertical and horizontal bars correspond to the original model size and accuracy. (Exponential y-scale.)}
\label{fig:class_progr}
\end{figure*}

\subsection{Reconstruction error} \label{secexp:resample}

We employ a normalized summed squared error:
$$nSSE(\mathbf{w}, \mathbf{\tilde{w}}) = \frac{{\Vert \text{vec}\left(\mathbf{w}\right) - \text{vec}\left(\mathbf{\tilde{w}}\right) \Vert_2^2}}{{\Vert \text{vec}\left(\mathbf{w}\right)\Vert_2^2}}$$ 
to measure the reconstruction error.
We experiment with one of the $1\times1$ layers of trained ResNet-50. The nSSE measured for the baseline compression without reordering the weights shows linear degradation as the fraction of used DCT coefficients decreases, signifying that the energy is not concentrated in lower frequencies as in many natural signals. The reordering procedure results in achieving lower nSSE given the same coefficient budget. Reshaping the tensor and decreasing the number of vectors $g$, which are transformed by DCT, further improves the result. The lower the value of $g$ the fewer coefficients are needed for achieving low nSSE. We have experimented with $\ell_1$ and $\ell_2$ norms for selection of the initial point for reordering, to find that both worked comparably, thus we have adopted the $\ell_2$. For the distance $\mathcal{D}$ that drives the reordering procedure we have considered Euclidean, Manhattan and Cosine distance. Manhattan distance shows slightly inferior results to the Euclidean. 
Cosine distance proved to be the best for large dimensional vectors $g\geqslant32$, while for smaller values of $g$ a better nSSE is obtained using Euclidean distance, which we use in all the following experiments.

\subsection{Classification accuracy} 
\label{secexp:combine}
We now compress all  convolutional layers at once and treat the output layer as a $1\times1$ convolutional layer. The ordering operation employed in our compression requires the order vector $\mathbf{s}$ to be stored. Since small values of $g$ require storing very long order vectors (and thus effectively reduce the compression ratio), we study only $g\geqslant 4$. The quality of compression is measured via classification accuracy. Note that compression rates $r$ do not take into account the storage of the index vector $\mathbf{s}$, but results in tables and figures are reporting the number of parameters including~$\mathbf{s}$, unless stated otherwise.

\noindent{\bf ResNet-50 architecture.} We have tested the compression strategy on ResNet-50 model with $g \in \left\{4,8,16\right\}$ and $r \in \left\{2,4,8,16,32\right\}$ for uniform compression  as well as for progressive-g compression and hyperparameter $r'\in\left\{0.125,0.25,0.5,1,2\right\}$ for progressive-r compression. 
Fig.~\ref{fig:class_progr} shows how the classification accuracy evolves when the proposed compression rate increases. We omit the first layer (having only $~$10k parameters).
A model compressed with progressive-r strategy with $g=4$ and $r'=0.125$ spanning 60\% of the original model size loses less than 1\% of classification accuracy without any fine-tuning, while a model without weight reordering loses all of its performance even after a mild compression.
We establish that a minor amount of fine-tuning enables us to approach the original level of accuracy (achieved by the non-compressed ResNet-50) even after substantial weight compression in the DCT domain. 
The fine-tuning is performed for one epoch only on central crops of the resized training images without using any augmentation. The learning rate is set to $0.001\times$batch\_size$/256$.
The performance of the  progressive-r compression exceeds that of the uniform for all values of $g$.
As can be seen in Fig.~\ref{fig:class_progr}, one epoch fine-tuning of a model uniformly compressed with $g=4$, $r=8\times$ allows to  perform as well as the original model  with the advantage of having only 38\% of its size. Comparable accuracy is obtained with progressive-r strategy ($r'=1$) with a model that has only 32\% the original amount of parameters, while the model without any reordering similarly compressed drops the accuracy by more than 17\%. 
The drawback of the progressive-r  strategy is the large vector of indices used to reorder the weights that sets a lower bound to the maximal possible compression. Progressive-g strategy addresses this issue by shrinking this vector for large layers and thus allows for higher compression rates without excessively corrupting the early layers. If we extend the fine-tuning time to 20 epochs (decreasing the learning rate by 10 at epoch 15 and using standard augmentation), a severely pruned model can improve its performance by at least another percentage point, see Tab.~\ref{tab:state_of_art}.
Note that models outperforming the proposed approach for stronger compression levels~\cite{he2018amc,renda2019comparing}, see Fig.~\ref{fig:class_progr}, involve very substantial amount of fine-tuning of 50-300 epochs.

We now compare the proposed spectral domain pruning to the standard magnitude-based pruning using ${\ell}1$-norm.  Alg.~\ref{alg:compression} can be simply adjusted to perform pruning of tensors in standard bases bypassing the DCT transform. The elements of size $g$ are sorted according to their $\ell_1$ norms, keeping only a relevant fraction of those with largest norms. Tab.~\ref{tab:standard_pruning} clearly demonstrates the superiority of DCT domain-compression for preserving network performance. Minor fine-tuning of one epoch is only $2\%$ worse than original model at $32\times$ DCT compression of all $1\times1$ and $3\times3$ layers, while the $\ell1$-pruned model cannot recover any of the original performance. 

\begin{table}[!t]
    \centering
    \tabcolsep = .6mm
    \begin{tabular}{l||c|c|c|c|c||c|c|c|c|c}
        \hline
        \multirow{2}{*}{reordering} & \multicolumn{5}{c||}{compress} & \multicolumn{5}{c}{compress \& fine-tune} \\
        & 2$\times$ & 4$\times$ & 8$\times$ & 16$\times$ & 32$\times$ & 2$\times$ & 4$\times$ & 8$\times$ & 16$\times$ & 32$\times$ \\
        \hline
        $\ell_1$ reorder & 16.2 & 0.1 & 0.1 & 0.1 & 0.1 & 74.4 & 68.2 & 54.1 & 22.5 & 7.0 \\
        Alg.~\ref{alg:reorder}+DCT & 74.8 & 70.4 & 49.5 & 14.4 & 0.2 & 76.6 & 76.5 & 76.1 & 75.6 & 74.0 \\
        \hline
    \end{tabular}
    \caption{ Top-1 classification accuracy on ImageNet validation set of ResNet-50 compressed with $g=4$. Combination of reordering and DCT transform improves over magnitude-based pruning ($\ell_1$).}
    \label{tab:standard_pruning}
\end{table}

\noindent{\bf MobileNet-V2 architecture.}
We now investigate the capability of our method to compress models with compact design. MobileNet-V2 model~\cite{sandler2018mobilenetv2} sparsifies the weights by using depth-separable convolutions. A vast portion of its weights thus resides in its resampling layers (1$\times$1 convolutions) and the output layer. We focus only on these layers, leaving the first layer and the following 7 blocks intact. The one epoch fine-tuning is done with batch size 64 and learning-rate 0.0001. The progressive compression approach enabled us to compress the network into 42\% of its original size within 1\% accuracy drop (see Fig.~\ref{fig:class_progr}). We also observe that  up until 50\% parameter compression the top-5 accuracy remains above the original level.

\begin{table}[!t]
    \centering
    \tabcolsep = 1.1mm
    \begin{tabular}{l||c|ccc|ccc}
        \hline
        \multirow{2}{*}{metric} &\multirow{2}{*}{Orig.} & \multicolumn{3}{c}{$g=4$} & \multicolumn{3}{|c}{$g=8$} \\
        & & $r'$=$\sfrac{1}{8}$ & $r'$=$\sfrac{1}{4}$ & $r'$=1 & $r'$=$\sfrac{1}{2}$ & $r'$=1 & $r'$=1* \\
        \hline
        \#params & 25.6M & 15.4M & 12.0M & 8.2M & 6.5M & 5.0M & 5.0M \\
        \#trainable & 25.6M & 8.9M & 5.5M & 1.7M & 3.2M & 1.7M & 1.7M \\
        top1 \% $\uparrow$ & 76.15 & 76.69 & 76.53 & 76.16 & 75.21 & 73.02 & 74.77 \\
        top5 \% $\uparrow$ & 92.87 & 93.27 & 93.20 & 92.96 & 92.36 & 91.37 & 92.27 \\
        \hline
    \end{tabular}
    \caption{Our progressive-$r$ compressed ResNet-50~\cite{he2016deep,Torchvision} results on ImageNet after one epoch of fine-tuning post-compression (*model is fine-tuned 20 epochs).}
    \label{tab:state_of_art}
\end{table}

\section{Conclusion}
\label{secconcl}

We have proposed a novel DCT-based compression approach for effectively reducing the CNN parameter footprint that relies on tensor reordering and pruning. Considering very limited fine-tuning that we employ in this work, our method demonstrates state-of-the-art performance as applied to ResNet-50~\cite{he2016deep} and MobileNet-V2~\cite{sandler2018mobilenetv2} architectures for image classification task on ImageNet dataset. Our model has also the capability to produce strongly performing compressed networks at zero-retraining.
Note  that quantization and Huffman coding can be used to yield memory efficient representation of our results. Future work will explore reducing computations as well as memory footprint using spectral representation.

\newpage
\clearpage
\bibliographystyle{IEEEbib}
\bibliography{biblio}

\end{document}